\documentclass{article}
\usepackage{spconf,amsmath,graphicx}
\usepackage[dvipsnames]{xcolor}
\usepackage{multirow}
\usepackage{hyperref}
\usepackage{booktabs} 
\usepackage{footmisc}

\title{Coarse-to-Fine Knowledge Selection for Document Grounded Dialogs}
\name{Yeqin Zhang$^{1,2*}$, Haomin Fu$^{1,2}$\sthanks{Equal contribution.}, Cheng Fu$^2$, Haiyang Yu$^2$, Yongbin Li$^{2\dag}$, Cam-Tu Nguyen$^1$\sthanks{Corresponding authors.}}
\address{$^1$State Key Laboratory for Novel Software Technology, Nanjing University, China\\ $^2$Alibaba Group\\ \{zhangyeqin, haominfu\}@smail.nju.edu.cn\\ \{fucheng.fuc, yifei.yhy, shuide.lyb\}@alibaba-inc.com, ncamtu@nju.edu.cn 
}
\newcommand\prompt[1]{$<$\textit{#1}$>$}

\begin{document}
%
\maketitle

\begin{abstract}
Multi-document grounded dialogue systems (DGDS) belong to a class of conversational agents that answer users' requests by finding supporting knowledge from a collection of documents. Most previous studies aim to improve the knowledge retrieval model or propose more effective ways to incorporate external knowledge into a parametric generation model. These methods, however, focus on retrieving knowledge from mono-granularity language units (e.g. passages, sentences, or spans in documents), which is not enough to effectively and efficiently capture precise knowledge in long documents. This paper proposes Re3G, which aims to optimize both coarse-grained knowledge retrieval and fine-grained knowledge extraction in a unified framework. Specifically, the former efficiently finds relevant passages in a retrieval-and-reranking process, whereas the latter effectively extracts finer-grain spans within those passages to incorporate into a parametric answer generation model (BART, T5). Experiments on DialDoc Shared Task demonstrate the effectiveness of our method.

\end{abstract}
\section{Introduction}
\label{sec:intro}
This paper targets DGDS, a class of knowledge-grounded conversational agents that obtain information from a collection of documents to answer users' requests. Document-grounded dialogue differs from open-domain question answering (open-domain QA)\cite{fu2020survey},  a closely related but well-studied problem, in two aspects. Firstly, DGDS are multi-turn systems, that is,  user requests in a (conversational) session are interconnected. Consequently, it is desirable for DGDS to effectively model the rich interaction between conversational histories and documents. Secondly, DGDS are closed-domain dialog systems, that is, the supporting documents belong to a narrow domain, e.g. manual books of a tech company. As a result, texts from such documents lie more closely in the semantic space, and thus a finer matching is needed to find suitable knowledge from typically long documents in DGDS.

Despite the difference, many methods in open-domain QA and open-domain dialogue systems \cite{dai2022cgodial, mi2021towards} (a multi-turn variant of open-domain QA) can be readily applied to DGDS. In particular, retrieval-and-generation (RAG) \cite{patrick2020rag} and retrieval-reranking-generation (Re2G) \cite{glass-etal-2022-re2g} have been exploited for DGDS with certain success \cite{feng2021multidoc2dial, fu2022doc2bot}. This line of research takes the advantage of nonparametric memories (of retrieved knowledge) to large parametric transformers \cite{colin2020t5, he2022galaxy}. A number of techniques can be applied to improve knowledge retrieval such as new training strategies \cite{yingqi2021rocketqa, liu2021dialoguecse, liu2022dial2vec}, and response-aware knowledge retrieval models \cite{chen-etal-2020-bridging,meng2020dukenet,kim2019sequential,lian2019learning,zhao-etal-2020-knowledge-grounded,zheng-etal-2020-difference}. Different strategies are also available to fuse knowledge into generation models \cite{patrick2020rag,gao-etal-2022-unigdd,izacard-grave-2021-leveraging,  li-etal-2019-incremental,li-etal-2019-incremental,meng2020refnet,lin2020generating, 9746877, xu2021few}. One can also get inspiration from open-domain dialogue systems to capture richer interactions between dialog histories and documents by considering multiple granularities of dialogue contexts \cite{tao-etal-2019-one, yuan-etal-2019-multi,gu2019interactive}. Unfortunately, these methods do not fully take into account the characteristics of DGDS. Specifically, prominent models only focus on retrieving mono-granularity language units (passages, sentences, or spans) from a document collection. 

\begin{figure*}
    \centering
    \includegraphics[width=0.83\textwidth]{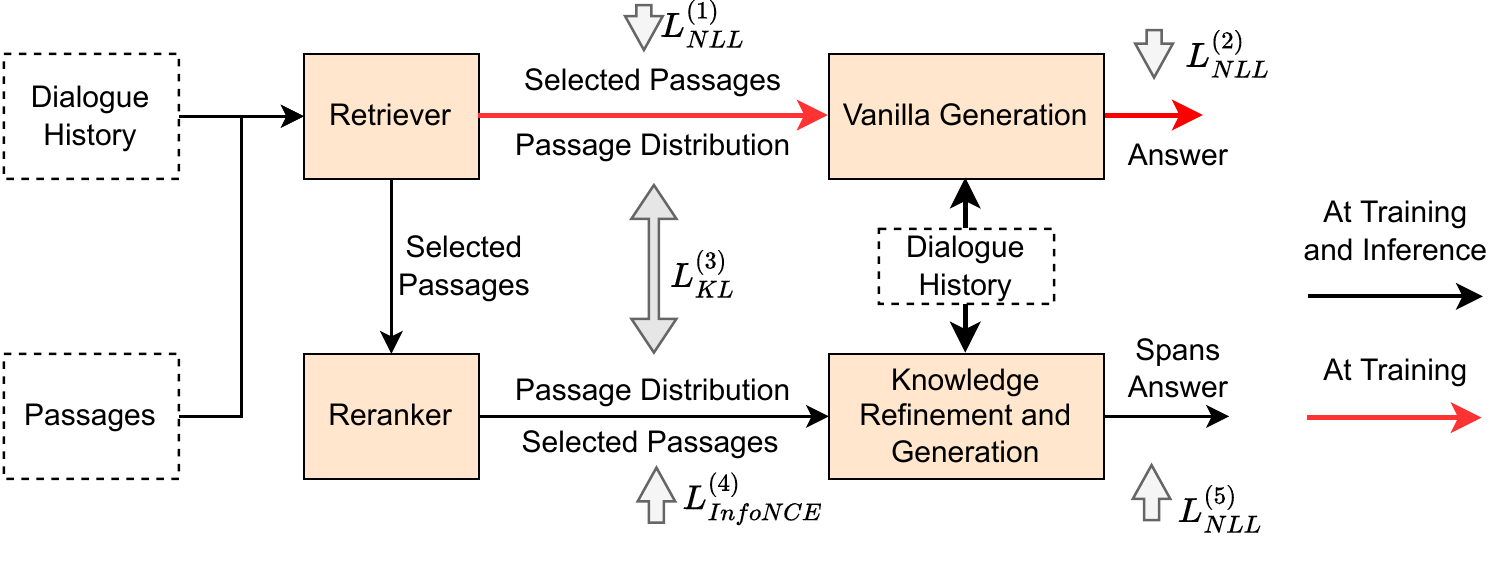}
    \caption{Re3G framework consists of three main components: (1) Retriever, (2) Reranker, (3) a joint model of Knowledge Refinement and Generation. Here, Vanilla Generation is used for finetuning the Retriever only.}
    \label{fig:re3g-architecture}
\end{figure*}

Recently, \cite{feng2021multidoc2dial} has considered knowledge span extraction (fine-grained knowledge selection) as an essential task for DGDS. Unfortunately, the study \cite{feng2021multidoc2dial} has not modeled span extraction jointly with passage retrieval for response generation. And thus, it is unclear how much fine-grained knowledge selection influences the final performance of DGDS. To bridge this gap, this paper proposes Re3G, which models coarse-grained (passage) retrieval, fine-grained (span) retrieval, and generation in a unified framework.  Specifically, the former efficiently finds relevant passages in a retrieval-and-reranking process, whereas the latter effectively extracts finer-grain spans within those passages to incorporate into a parametric answer generation model (BART, T5). We conduct thorough experiments on DialDoc Shared Task \cite{feng2021multidoc2dial}, and show that both coarse-grained and fine-grained knowledge selection are essential for response generation in DGDS.

\section{Method}\label{sec:method}

\subsection{Overview}

The problem of DGDS can be described as follows. Assuming a collection of $D$ documents, we first split each document into multiple passages and get $M$ total passages $\mathcal{D}=\{p_1,...,p_M\}$. Given a conversational history $C=(u_1,a_1,...,u_{t-1},a_{t-1},u_t)$, where $u_i$ and $a_i$ are respectively the i-th user and agent utterances, our task is to generate the next system answer $a_t$ based on the passage collection.  

Knowledge selection is essential for open-domain QA as well as DGDS. Most of prior studies, however, do not simultaneously consider different levels of granularity of knowledge. For example, \cite{karpukhin2020dpr, patrick2020rag, glass-etal-2022-re2g, yingqi2021rocketqa,feng2021multidoc2dial} only focus on passage retrieval from a large collection of passages, whereas  \cite{chen-etal-2020-bridging,meng2020dukenet,kim2019sequential,lian2019learning, zhao-etal-2020-knowledge-grounded, zheng-etal-2020-difference, gao-etal-2022-unigdd, feng2020doc2dial} aim to select sentences or spans from a smaller pool of sentences. Towards a practical DGDS, Re3G aims at optimizing both course-grained knowledge retrieval (the Retriever and the Reranker) and fine-grained knowledge extraction (the Refinement component) in a unified framework. The general framework of Re3G is given in Figure \ref{fig:re3g-architecture}, and the details are described in the following.

\subsection{Model Details}
\subsubsection{Passage Retrieval}
The goal of the Retriever model is to efficiently find the top $k$ ($k<<M$) passages relevant to a dialogue context. Following \cite{karpukhin2020dpr, patrick2020rag, glass-etal-2022-re2g}, we apply the Bi-encoder architecture \cite{10.1145/2505515.2505665, reimers-gurevych-2019-sentence} for its efficiency. Specifically, a dialogue context $C$ and a passage candidate $p$ are first encoded by two separate BERT (base) encoders  \cite{jacob2019bert}. The similarity scoring function is then defined as the dot product of the encoder outputs:
\begin{equation*}
    \begin{split}
    P_{ret}(p|C)&\propto dot[BERT_c(C), BERT_p(p)]\\
    &=sim_{\eta_1}(C,p)
    \end{split}
\end{equation*}

\noindent \textbf{Training} In order to improve the effectiveness of the Retriever while keeping its efficient architecture, we follow Re2G\cite{glass-etal-2022-re2g} to train the Retriever in three phases as follows.

In the \textit{first phase}, given a training set of $\{(C, p_i^+\in \mathcal{P}^+,p^-_j\in \mathcal{P}^-)\}$, where $\mathcal{P}^+$ and $\mathcal{P}^-$ indicate the set of relevant (positive) and non-relevant (negative) passages for the context $C$, we train the query and the passage encoders by minimizing the sum of negative log likelihoods $L^{(1)}_{NLL}$.

In the \textit{second phase}, we finetune the context encoder with guidance from a response generation like RAG \cite{patrick2020rag}  using negative log-likelihood ($L^{(2)}_{NLL}$ in Figure \ref{fig:re3g-architecture}). This finetuning process is beneficial for companies that own a large number of ungrounded conversational logs. Unlike RAG \cite{patrick2020rag}, this generation model, which is referred to as Vanilla Generation hereafter, is not used for response generation at inference. This design helps simplify the training process of Re3G by decoupling the Retrieval component and the Refinement/Generation component (of Section \ref{sec:knowledge-generation}).

In the \textit{third phase}, we further finetune the Retriever with the guidance of the Reranker model (Section \ref{sec:reranking}) using the teacher-student method like \cite{glass-etal-2022-re2g}. Specifically, we update the context and passage encoders to minimize the KL divergence ($L^{(3)}_{KL}$) between the passage distributions calculated using the Retriever and the Reranker model.



\subsubsection{Passage Reranking} \label{sec:reranking}
Given a shortlist of candidates, the goal of a Reranker is to capture deeper interactions between a dialogue context and a candidate passage. Specifically,  the passage and the context are concatenated to form the input for RoBERTa\cite{yinhan2019roberta}. A linear layer and sigmoid layer are used on top of the [CLS] output of the RoBERTa model to obtain the similarity.
\begin{equation*}
\begin{split}
    P_{rerank}(p|C)&\propto\text{sigmoid}\{\text{linear}[RoBERTa([p,C])]\}\\
    &=sim_{\eta_2}(p,C)
\end{split}
\end{equation*}

\noindent \textbf{Training} Compared to the Retriever, the Reranker is expected to be more effective with highly relevant passages at lower rank positions, hence reducing the risk of excluding true positive (relevant) passages or introducing noises from false positives. As a result, InfoNCE loss is used to train our Reranker.
\begin{equation*}
    L^{(4)}_{InfoNCE} = -\log \frac{\exp(sim_{\eta_2}(C, p^{+})/\tau)}{\sum_{p\in {\mathcal{P}}^{\pm}}\exp(sim_{\eta_2}(C, p)/\tau)}
\end{equation*}
where $\tau$ is a temperature hyper-parameter. A small value of $\tau$ drives the distribution over candidate passages closer to one-hot vector, effectively encouraging a small number of positive passages to be separated from the abundance of irrelevant (negative) ones. In our study, we empirically find the value of $\tau$ of 0.07 to be effective for our problem.



Negative sampling has been found to be essential for training dense retrieval models \cite{yingqi2021rocketqa,DBLP:conf/iclr/XiongXLTLBAO21}. We apply a simple yet effective sampling strategy, in which we randomly sample 30 negative samples from a pool of around 100 top passages obtained from the Retriever model. This sampling process is done in every training epoch of the Reranker, resulting in a dynamic negative sampling strategy. 



\subsubsection{Knowledge Refinement and Generation} \label{sec:knowledge-generation}
Given the dialogue context $C$ and top passages $\{p_1,\ldots,p_5\}$ from the Reranker, we aim to generate knowledge grounding and answer $(z,a)$. In general, many finer-grain knowledge extraction architectures \cite{chen-etal-2020-bridging,meng2020dukenet,kim2019sequential,lian2019learning, zhao-etal-2020-knowledge-grounded, zheng-etal-2020-difference, gao-etal-2022-unigdd} and text generation methods \cite{patrick2020rag, izacard-grave-2021-leveraging,9413753, 9747600,  gao-etal-2022-unigdd,li-etal-2019-incremental,li-etal-2019-incremental,meng2020refnet,lin2020generating} can be used for knowledge refinement and generation. For simplicity, we apply a prompt-based multi-task learning method similar to \cite{gao-etal-2022-unigdd} for jointly modeling the knowledge refinement and generation. Note that, however, \cite{gao-etal-2022-unigdd} works by fixing the document in advance, while we need to extract finer grained knowledge from retrieved passages of the Reranker.

Following \cite{gao-etal-2022-unigdd}, we train a T5 sequence-to-sequence generation model \cite{colin2020t5} on carefully constructed pairs of inputs and outputs. The rule for constructing the input is ``[task template] [history template] [passage template]''. Here, ``[task template]'' can be one of the following: 1) ``generate \prompt{ground} then \prompt{agent}'' for both knowledge refinement and response generation; 2) ``generate 
\prompt{ground}'' for only knowledge refinement; and 3) ``generate \prompt{agent}'' for answer generation. The ``[history template]'' is encoded by a sequence of ``\prompt{user} [user utterance] \prompt{agent} [agent utterance] ...'', where utterances are sentences from the dialog context $C$. The ``[passage template]'' is obtained simply by concatenating top relevant passages ${p_1,...,p_5}$ from the Reranker. During training, the ground truth output is constructed depending on the ``[task template]''. For example, if we obtain the input with prompt ``generate \prompt{ground} then \prompt{agent}" then the corresponding output is built by ``\prompt{grounding} [groundtruth span] \prompt{agent} [groundtruth answer]''. By doing so, we obtain a sequence-to-sequence training dataset, which connects the knowledge refinement and generation tasks by special prompts. During inference, the input is prepared similar to the training phase, and the output is then parsed to obtain extracted span, which follows ``\prompt{grounding}'' prompt, and the answer, which follows ``\prompt{agent}'' prompt. 



\noindent \textbf{Training} The joint model of knowledge refinement task and generation task is trained based on ground truth spans and responses using negative log-likelihood loss $L^{(5)}_{NLL}$. In order to effectively train both tasks, the training process is conducted in two stages. Specifically, both tasks are trained jointly in the first stage, we then further finetune each task for one more epoch in the second training phase.

\section{Experiments}
\label{sec:experiments}

\subsection{Evaluation Dataset and Metrics}
\label{sec:experiment-settings}

We conduct our experiments on DialDoc Shared Task\footnote{\url{https://doc2dial.github.io/multidoc2dial}}, which is an extension of the Multidoc2dial dataset \cite{feng2021multidoc2dial}. For training, we are provided with a training set of 3,474 dialogs corresponding to 48,002 utterances, and a validation set of 661 dialogs corresponding to 9,195 utterances. Each utterance  is annotated with a number of grounding passages and spans. For testing, we use two test sets from Seen Leaderboard, which are referred to as DevTest with 199 dialogs and Test with 661 dialogs. The inferred system responses on DevTest and Test sets are then uploaded to the Leaderboard for evaluation using token-level F1 score (F1), SacreBLUE (S-BLEU), and ROUGE-L \cite{feng2021multidoc2dial}. Note that we focus on Seen Leaderboard to study the effectiveness of the reranking and refinement components in a closed-domain setting, which is a practical setting for DGDS. 



\begin{table*}
    \centering
    \begin{tabular}{l|ccc|ccc}
        \toprule
        \multirow{2}*{Method} & \multicolumn{3}{c|}{DevTest} & \multicolumn{3}{c}{Test}\\ \cmidrule{2-4} \cmidrule{5-7} 
         & \textbf{F1} & \textbf{S-BLEU} & \textbf{ROUGE-L}& \textbf{F1} & \textbf{S-BLEU} & \textbf{ROUGE-L}\\ \midrule
        RAG-BART$_{large}$  & 36.23 & 21.41 & 34.01 & 35.85 & 22.26 & 33.82 \\
        Re2G-BART$_{large}$ & 39.18 & 24.19 & 36.66 & 40.16 & 27.49 & 38.43\\
        Re2FiD-T5$_{large}$ & 45.43 & 31.32 & 43.11 & 46.62 & 32.84 & 44.87\\
        Re3FiD-T5$_{large}$ & 46.71 & \textbf{33.52} & 44.71 & 46.77 & 33.48 & 45.12 \\
        Re3G-T5$_{large}$ & 48.25\ddag  & 31.44 & 45.81 \ddag & 49.33\ddag & 34.67\ddag & 47.34 \ddag\\
         { } { } { }-- Refinement & 47.93 & 30.73 & 45.41 & 49.01 & 33.54 & 47.04 \\
         { } { } { }-- Rerank  & 45.76 & 31.08 & 43.12 & 45.91 & 31.46 & 44.06 \\\midrule
        Re3G-T5$_{3b}$ & \textbf{49.05} & 32.12\ddag & \textbf{46.34} & \textbf{50.38} & \textbf{35.90} & \textbf{48.44}\\ 
         { } { } { }-- Refinement& 45.49 & 31.27 & 43.07 & 47.92 & 35.03 & 45.99\\
         { } { } { }-- Rerank & 45.92 & 31.12 & 43.14 & 46.80 & 32.78 & 44.55 \\
        \bottomrule
    \end{tabular}
    \caption{Results of compared methods on Seen Leaderboard of DialDoc Shared Task. The \textbf{best} and runner-up{\ddag} are marked.}
    \label{tab:seen-results}
\end{table*}

\subsection{Models for Comparison} 
\textbf{RAG} \cite{patrick2020rag} follows the Retrieval-and-Generation architecture, where we use BERT-base\footnote{\url{https://huggingface.co}\label{fn:1}} for retrieval and BART-large\footref{fn:1} for text generation. For knowledge-grounded generation, RAG introduces relevance bias into text generation at the token level. By relevance bias, we mean the higher relevant passage has more influence on the response generation.

\noindent\textbf{Re2G} \cite{glass-etal-2022-re2g} follows the Retrieval-Rerank-Generation architecture. Here, we use the same setting as in \cite{glass-etal-2022-re2g} for experiment. Specifically, we use BERT-base for retrieval, another BERT-base for reranking, and BART-large for generation. During inference, 24 candidates from the retrieval are used for reranking. The generation is based on relevance bias like RAG.

\noindent\textbf{Re2FiD} is a modification of Re2G, where the retriever and the reranker are the same as our method (Re3G). More specifically, we use a BERT-base for retrieval, RoBERTa\footref{fn:1} for reranking, BART-large for vanilla generation, and T5-large for generation. Unlike RAG and Re2G, the generator of Re2FiD follows Fusion-in-Decoder (FiD) approach \cite{izacard-grave-2021-leveraging}, which encodes the top 5 passages and context independently before concatenation for the decoder. During inference, 100 candidates from the retrieval are used for reranking, which is the same as Re3FiD and Re3G described in the following.

\noindent\textbf{Re3FiD} adapts Re2FiD to take knowledge refinement into account. The retriever and reranker are the same as Re2FiD. However, Re2Fid uses a prompt-based multi-task design like R3G. Unlike Re3G, prompts, top passages, and contexts are encoded independently using the FiD approach.

\noindent\textbf{Re3G} is the same with Re3FiD except that we use the early concatenation instead of FiD. Note that early concatenation allows richer interactions via self-attention compared to FiD. Two variants, Re3G-T5$_{large}$ and Re3G-T5$_{3b}$, are tested to study the impact of 2 pre-trained models, T5-large\footref{fn:1} with 700 million parameters and T5-3b\footref{fn:1} with 3 billion parameters. 



\subsection{Experimental Results and Discussion}
We report our experimental results in Table \ref{tab:seen-results}, where new findings are summarized as follows:

(1) The quality of coarse-grained (passage) retrieval and fine-grained (span) retrieval are both essential for DGDS. By adding a reranking component, the performance of passage retrieval is improved, resulting in an improvement in Re2G-BART$_{large}$ compared to RAG-BART$_{large}$. This is consistent with the finding in \cite{glass-etal-2022-re2g}. On the other hand, the integration of the knowledge refinement component for fine-grained retrieval contributes to a significant improvement of Re3FiD-T5$_{large}$ over Re2FiD-T5$_{large}$. The essential roles of the reranking and the knowledge refinement can also be seen by comparing Re3G to Re3G variants without either of these two components. Table \ref{tab:seen-results} shows that removing either of these components results in a drop in the performance of Re3G. 

(2) Knowledge refinement in Re3G is better trained with a large pre-trained model. Since we use the prompt-based method for the knowledge-refinement component, a larger pre-trained model helps improve the quality of knowledge refinement, subsequently leading to better performance for DGDS. As observable from Table \ref{tab:seen-results}, the impact of the knowledge refinement is larger when we use T5$_{3b}$ instead of T5$_{large}$. Specifically, Re3G-T5$_{3b}$ decreases by 3.55 in F1 when we remove the knowledge refinement component. In contrast, Re3G-T5$_{large}$ only decreases by 0.32 in F1 when we remove the knowledge refinement component.

(3) It is recommended to use early fusion instead of FiD to incorporate grounding knowledge into text generation. Comparing Re3G-T5$_{large}$ to Re3FiD-T5$_{large}$, the only difference is that Re3G-T5$_{large}$ exploits early fusion instead of FiD. As we can see from Figure \ref{tab:seen-results}, the early fusion strategy results in a better performance for Re3G-T5$_{large}$ in most of the evaluation metrics. This is reasonable given the fact that early fusion allows richer interaction between knowledge and contexts compared to the FiD approach.

It is also noteworthy that Re3G is comparable to SOTA on the Leaderboard even without ensemble trick.

\section{Conclusion}
This paper studies the effectiveness of coarse-to-fine grained knowledge selection on DGDS. We propose a novel method named Re3G, that optimizes the coarse-grained (passage) retrieval via a reranking component, and the fine-grained (span) retrieval via a knowledge refinement component.  Additionally, we propose an efficient training process, where we make two technical contributions: 1) a vanilla generation is introduced to efficiently train the retriever; and 2) a dynamic negative sampling for training the reranker. Experiments on DialDoc Shared Task show that both coarse and fine-grained knowledge selection are essential for DGDS. Our experiments also suggest that: 1) a large pre-trained model should be used to improve the quality of both knowledge refinement and generation in Re3G; and 2) early fusion is essential to model the rich interaction between knowledge and contexts.

\bibliographystyle{IEEEbib}
\bibliography{main}

\end{document}